# 3D Printer-Controlled Syringe Pumps for Dual, Active, Regulable and Simultaneous Dispensing of Reagents. Manufacturing of Immunochromatographic test strips.

Gabriel Siano*[a,b], Leandro Peretti[c,d], Juan Manuel Márquez[b], Nazarena Pujato[b,] Leonardo Giovanini[a,e], Claudio Berli[c]

(a) Instituto de Investigación en Señales, Sistemas e Inteligencia Computacional (sinc(i), UNL-CONICET), Argentina
(b) Facultad de Bioquímica y Ciencias Biológicas, Universidad Nacional del Litoral, Argentina
(c) Instituto de Desarrollo Tecnológico para la Industria Química (INTEC, UNL-CONICET), Argentina
(d) Facultad de Ciencias Médicas, Universidad Nacional del Litoral, Argentina.
(e) Facultad de Ingeniería y Ciencias Hídricas, Universidad Nacional del Litoral, Argentina

* Corresponding author: gsiano@sinc.unl.edu.ar (G. Siano), Tel: +54 0342 4575233 (117)
Postal address: Ciudad Universitaria UNL, Ruta Nacional N° 168, km 472.4, FICH, 4to Piso (3000) Santa Fe – Argentina

## Abstract:

Lateral flow immunoassays (LFIA) are widely used worldwide for the detection of different analytes because they combine multiple advantages such as low production cost, simplicity, and portability, which allows biomarkers detection without requiring infrastructure or highly trained personnel. Here we propose to provide solutions to the manufacturing process of LFIA at laboratory-scale, particularly to the controlled and active dispensing of the reagents in the form the Test Lines (TL) and the Control Lines (CL). To accomplish this task, we adapted a 3D printer to also control Syringe Pumps (SP), since the proposed adaptation of a 3D printer is easy, free and many laboratories already have it in their infrastructure. In turn, the standard function of the 3D printer can be easily restored by disconnecting the SPs and reconnecting the extruder. Additionally, the unified control of the 3D printer enables dual, active, regulable and simultaneous dispensing, four features that are typically found only in certain high-cost commercial equipment. With the proposed setup, the challenge of dispensing simultaneously at least 2 lines (CL and TL) with SPs controlled by a 3D printer was addressed, including regulation in the width of dispensed lines within experimental limits. Also, the construction of a LFIA for the detection of leptospirosis is shown as a practical example of automatized reagent dispensing.

# 1. Introduction

Lateral flow immunoassays (LFIA) are widely used worldwide for the detection of different analytes because they combine multiple advantages, such as low production cost, simplicity, and portability, which allows biomarkers detection without requiring infrastructure or highly trained personnel. This kind of immunochromatographic technique has been used for the detection of various pathological agents such as viruses, bacteria, fungi [1], toxins[2], antibiotics[3], pesticides[4], herbicides[5], among others.

The manufacture of LFIA involve: i) the synthesis and characterization of colloidal gold particles with sizes close to 40 nm, ii) their subsequent sensitization with biomolecules (antigens or specific antibodies), iii) the dispensing of the gold-biomolecules complexes on specific supports, iv) the dispensing of reagents in the form of lines on nitrocellulose membranes (NCM), which will give rise to the biorecognition reactions in areas called Test Line (TL) and Control Line (CL), and finally v) the assembly of the different parts (sample pad, absorbent pad, housing) to obtain the test strip.

In laboratory prototypes, TL and CL are usually made manually, in order to evaluate different variables such as concentration and volume of reagents. However, a subsequent step involves automating reagent dispensing to achieve good reproducibility of the system. Although there is commercial equipment dedicated to resolving this aspect, it is also possible to develop cheaper alternatives that may be within the reach of more laboratories. On the other hand, commercial equipment, especially those based on Syringe Pumps (SP), usually have a significant dead volume, which is not relevant when a large number of IFLs are produced. However, in the development stages of a new prototype it is important to minimize reagent losses. In addition, when a prototype presents good results, the automation of the dispensing of the reagents on the NCM allows increasing the production of IFLs. The reproducible production of a significant number of LFIA is a key step to validate the devices in front of a high number of real samples and to reliably determine their diagnostic parameters, such as sensitivity and specificity.

Here we propose to provide solutions to the manufacturing process of LFIA for laboratory-scale, particularly to the controlled dispensing of the reagents in the form of lines. To accomplish this task, we adapted a 3D printer to also control SPs, since the adaptation of a 3D printer is economical and many laboratories already have it in their infrastructure.

An FFF (Fused Filament Fabrication) 3D printer for conventional thermoplastics (PLA, PETG, ABS, etc) can be made or can be bought with a low budget. This kind of 3D printers exist massively and a lot of them are open source (hardware, firmware and software). The maker community has achieved a high level of documentation and/or exemplification. All these characteristics make 3D printers instruments with a high level of accessibility.

At the same time, 3D printers are usually self-replicant (the 3D printer can print parts of itself) and the addition of designed pieces on or near the actuator (that is, the piece/carriage that holds the hotend) can be achieved within certain weight limits. Recently, a study[6] was reported in which a technical pen was attached to the carriage of a 3D printer. This setup is restricted to passive dispensing only, and due to the rigidity of the DT, minor imperfections can cause the NCM to become scratched. Given the limitation to passive dispensing, multiple-pass experiments on the NCM were reported. In the current study, however, the dispensed quantity is under active and configurable control, and optionally, it is also possible to support two (even more) technical pens or other DTs for passive dispensing.

Beyond merely utilizing the 3D positioning capabilities to move a dispensing tool for active or passive dispensing, some actions can be carried out with simple G-code, such as controlling the temperature, level and speed of the printing bed, and also general handling of steppers motors (axes, extruders, etc). This implies that the instrument can be modified and can easily drive stepper motors, which relates it to the control of steppers-based SP. This kind of SP also exists in 3D-printable and open source models[7–9], as the ones used in the present work, and can be used to perform different analytical techniques, such as lab-in-syringe[10,11]. Alternatively, a 3D printing kit was transformed into a programmable syringe pump set[12], controlled by the same firmware as in the present work (Marlin[13]), but using the respective connections to the X, Y, and Z axes. Being able to implement G-code avoids the need for programming the pumps from scratch, which is an advantage. Similarly, with other firmwares[14], it has been possible to control multiple SPs simultaneously. However, in both cases, the 3D movement axes were used, without preserving the standard function of a 3D printer, which was indeed one of the objectives of the present study.

Commercial equipment for LFIA manufacturing, which may or may not have active dispensing, is usually expensive, limited to working in a single dimension, and usually has no other utility. On the other hand, some devices or setups have been reported in the literature. One example is an open source 3D printable antibody dispenser[15], which works with a continuous NCM and requires absorption by the membrane. That is, the reagent is dispensed passively by the wicking force caused by the pores of the NCM[15] and/or by gravity. It should be emphasized that active dispensing allows for precise control of the liquid quantity. There are no gravity-related side effects, as liquid movement is controlled through pressure/vacuum. In contrast, passive dispensing may lead to results that cannot be optimized or may vary when changing the absorption properties of the NCM. In another related research[16], active dispensing with SPs resulted in more consistent lines, compared to flow driven by both hydrostatic pressure and wicking force. In that work two commercial SPs were used, one to pump liquid and the other modified as a linear actuator to move the Dispensing Tip (DT). However, this methodology is limited to working in a single dimension, and if more than one line is required, they must be obtained sequentially and not simultaneously. Since these tests require at least 2 lines (Test and Control lines) of different composition, and taking into account that, in addition to the challenge of handling the typically limited available volumes, the liquid replacement step in the DT would be added, simultaneity becomes essential. In turn, a unified control for both SPs was not reported, since the one used to move the DT had independent control, so both SPs must be manually coordinated by the operator. Since this adds variability to the system, it was avoided in our work. Both SPs and NCM movements were controlled by the 3D printer. Recently, another syringe-based Arduino-operated antibody dispenser has been reported[17]. The proposed setup is configured directly through hardware (potentiometers) but does not use G-code. It is restricted to dispensing a single line and imperfections were reported attributed to instability in the stepper motor-driven conveyor belt used.

In this study, certain aspects of the proposed dispensing procedure were examined, particularly with regard to linewidths. Firstly, the challenge of dispensing not sequentially but simultaneously at least 2 lines (control and test) with SPs controlled by a 3D printer was addressed. Since it may be desirable to obtain two different linewidths for TL and CL (considering signal intensity, costs, among other factors), and this requires independence between the lines in addition to simultaneity, active and passive methods of achieving this are discussed. Additionally, the construction of a LFIA for the detection of leptospirosis is presented as a practical application. Leptospirosis is a globally distributed zoonotic disease, prevalent in both urban and rural areas, and is caused by spirochetes belonging to the genus *Leptospira spp*[18]. Recently, a laboratory prototype of an LFIA was developed for the detection of leptospirosis using antigens derived from Leptospira strains circulating within Argentina[19]. In this work, LFIA devices for the detection of human leptospirosis were obtained and assessed using positive and negative control samples.

# 2. Material and Methods

2.1 Instruments and materials

A modified Hellbot Magna I 3D printer was utilized. The modifications involved the addition of wires and an A4988 stepper driver for an extra stepper/extruder, a firmware change, and the attachment of a 3D-printed Dispensing Tips Holder to the printer's carriage. The firmware was updated to Marlin 2[13] and the Mixing Extruder mode[20] was enabled. Pronterface was utilized to control the 3D printer and send runtime G-code. It should be noted that once a setup has been optimized and a debugged G-code is available, it is also possible to save it on a memory card and dispense directly without the need for a PC.

For the assembly of the Syringe Pumps, PLA was used for the printed parts along with all the essential components required for the assembly of 2 SPs[7,9]. An 8 mm lead/2 mm pitch screw (very usual in Z-axis of 3D printers) was utilized as leadscrew.

Smiths medical 22G Jelco® IV Catheter Radiopaque were used as Dispensing Tips.

SPs and DTs were coupled with 3-Way Valves (3WVs), PTFE tubing (0.8 mm id) and 21G needles.

Printex® 3WVs were utilized, with 1 male and 2 female Luer connectors. When necessary, male-male Luer adapters were obtained by connecting two male connectors from Secondary Medication Sets (Euromix®) using the included tubing.

To load dispensing liquids into the DTs, standard 1 mL syringes were employed, while standard 10 mL syringes were used for accumulating displacement fluid.

Bovine serum albumin (BSA, Sigma-Aldrich), food dyes and DDI water were employed to obtain coloured solutions.

2.2 Edge Detection and Image Analysis:

In order to determine the width and uniformity of the lines dispensed through image analysis of different NCM, Canny[21] edge detection was employed, implemented in Python using OpenCV[22]. Two edges were detected for each line. These edges were analyzed along the entire path of the lines, and average linewidths in mm were obtained every 2.5 mm of linear movement.

2.3 Leptospira Antigen

Antigen was prepared as reported in the literature[23]. Briefly, a culture of *L. interrogans* Hardjo serovar was carried out until exponential phase and subsequently lysed by sonication. Phenyl-methyl-sulfonyl fluoride (PMSF, 1mM) was added to inhibit proteases and the concentration was determined by the method of Bicinchoninic Acid (BCA, Pierce Reagents)[24]. The lysate was stored at -20°C until use.

2.4 Preparation of gold-labeled anti-human IgG (a-IgG$_{hum}$/GNPs)

First, GNPs (~40 nm) were prepared[25]. Briefly, 100 ul of trisodium citrate solution were added (final concentration 0.015% w/v) to 50mL of AuCl3-·HCl·4H2O solution (0.01% w/v) under stirring at 100°C. After 30 minutes, the reaction is stopped and allowed to reach room temperature. GNPs size was determined by UV/vis spectrophotometer. Then, GNPs were employed to obtain the conjugate with Goat anti-human IgG (a-IgG$_{hum}$, Sigma)[25]. Passive sensitization was performed at pH=9 (borate buffer 20mM) by mixing for 30 min 10 mL of GNPs with 0.2 mg/mL a-IgG$_{hum}$ under stirring. Subsequently, the free sites of nanoparticles were blocked with BSA 1% final concentration. Excess of proteins were removed by centrifugation and redispersion cycles in 20 mM borate buffer pH=9, supplemented with 5% sucrose, 0.2% Tween-20, 0.2 M NaCl and 0.1% sodium azide, and the conjugate was stored at 4°C until use.

2.5 Assembly of the strip tests

TL and CL were deposited on the NCM (FF120HP, GE Life Sciences) as previously described, using a solution of 0.5 µg/µl of leptospiral antigen and 1 µg/µl of a rabbit anti-goat IgG (a-IgG$_{goat}$, Sigma-Aldrich), respectively. NCM were dried at 50°C for 30 min. The conjugate pad (Standard 17, GE Life Sciences) was embedded with the a-IgG$_{hum}$/GNPs solution previously prepared and dried at 50°C for 30 min. Membranes were assembled on a backing card according to standard descriptions, each pad overlapped 2 mm with each other, leaving the NCM below both conjugate and absorbent pads (CF3, Whatman, GE Life Sciences), and then the sample pad (CF3). Finally, membranes were cut by a paper cutter every 4 mm and placed in their housing.

2.6 Serum samples

Two sera previously classified by reference techniques into positive control and negative control were used, provided by the Laboratorio Nacional de Referencia de Leptospirosis (LNRL), Instituto Nacional de Enfermedades Respiratorias (INER), Santa Fe, Argentina. The use of these samples was approved by the Bioethics Committee of the Facultad de Bioquímica y Ciencias Biológicas of the Universidad Nacional del Litoral (CE2019-37, Acta 05/19).

2.7 Evaluation of the LFIA

Samples were diluted 1/10 in phosphate buffered saline (PBS) pH=7,6 and 100 µl were applied on the sample zone of the housing. At 15 minutes, the visualization of 2 lines (TL and CL) indicates a positive result, and the visualization of only de CL indicates a negative result.

# 3. Results and Discussion

### 3.1. Experimental setup and general considerations

The primary objective of this study was to perform a dual, active and simultaneous dispensing of two lines with different widths, which was achieved with the experimental setup depicted in figure 1. Control over the linewidths is sought, but obtaining a mathematical expression that describes them exactly depends on parameters beyond those of a 3D printer. NCM properties (wettability, absorption, hydrophilicity, pore size, flatness), as well as fluid properties (vapor pressure, concentration of reactive substances, viscosity), are undoubtedly significant. Therefore, the aim of this study is narrowed down to empirical control within certain experimental limits, while maximizing the highest degree of independence among widths. Certainly, commercial instruments also necessitate a previous empirical stage to evaluate the quality of the dispensed lines.

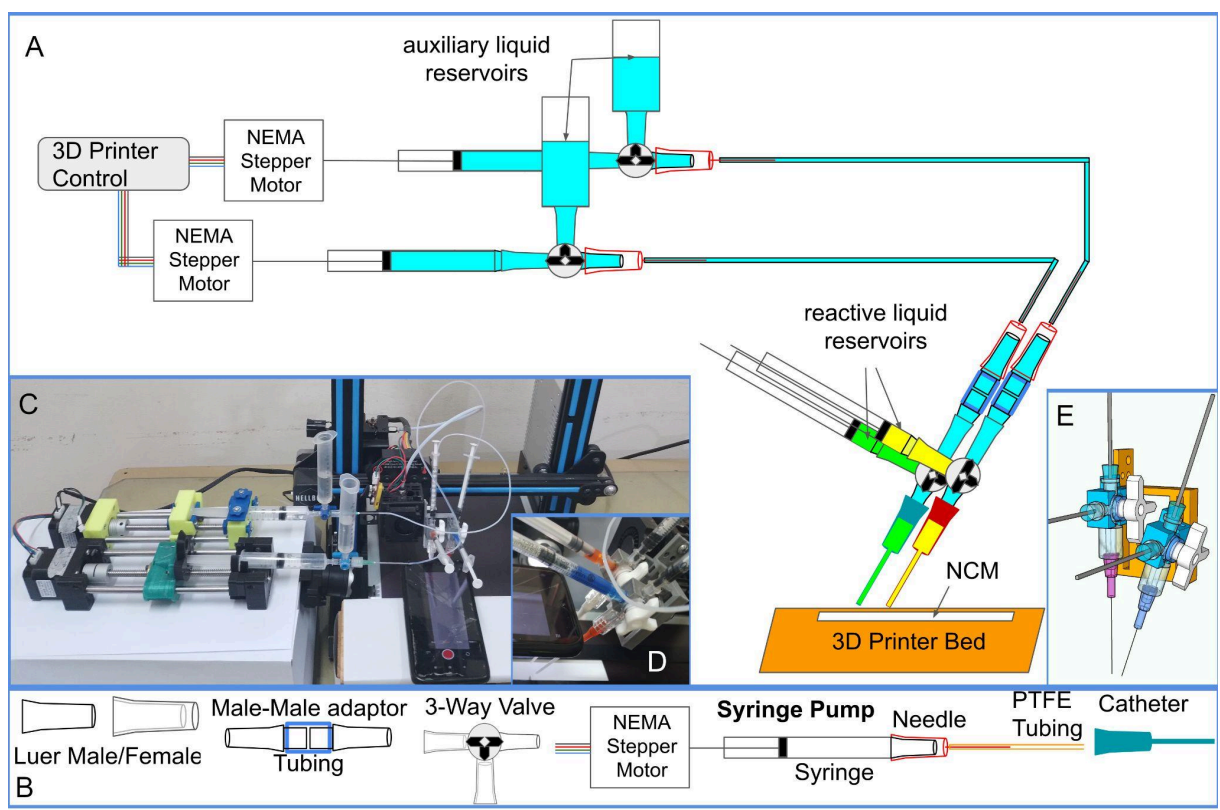

Figure 1: Experimental setup. A) Diagram of the experimental setup B) Symbols of components used in the experimental setup C) Photograph of the experimental setup D) Side view of tilted DTs dispensing onto an NCM E) Diagram of the dispensing end.

In figure 1 A, an outline of the experimental setup is depicted. Firstly, it is worth noting that the 3D printer has control over not only 2 SPs but also all the parameters typically controlled by a 3D printer, such as temperatures, positions, velocities, etc. To keep the 3D positioning functions, the connectors of the X, Y, and Z axes are not used to command SPs. In other words, one objective of this study was not to interfere with the normal operation of the 3D printer. Only in the case of dispensing, the plastic extruder will be disconnected, and one of the SPs will be connected in its place. In turn, an independent wiring system is schematized for a second SP. Since it is common for printers to have a single extruder, it will be necessary to install an additional power driver, along with an appropriate cable for another SP. After that, each SP will be able to work with a specific flow rate, achieving the highest degree of independence between linewidths. Once this has been accomplished, apart from a single firmware update to enable a second extruder (which will be discussed later), nothing more will be required other than connecting the original extruder to resume the normal plastic printing functions. In any case, the power drivers and the wiring to each SP must be adequate to transmit the necessary energy to the SPs, and any change in configurations, both for the 3D printer and the SPs, must be able to be done with G-code in runtime.

In the development of LFIA prototypes at laboratory scale, it is common to have a limited volume available for dispensing, sometimes in the order of microliters. This implies that any dead volume must be minimized. It also means that there will be little chance of pre-testing and that the previous experiments (calibration of speeds, distances, etc) must be carried out with liquids whose composition must be similar to that of the liquid of interest. In diagram A of figure 1, the previously mentioned small volumes are represented by syringes and catheters (DTs) containing green and yellow liquids. By utilizing 3WVs at the dispensing ends, it is possible to manually load these small volumes directly into the catheters. These valves are basic nursing/medical supplies, they allow for splitting of fluids and they can be coupled with other 3WVs (1 male and 2 female ends). Unlike passive dispensing systems, the described system incorporates SPs for active loading or unloading, making it possible to also perform loading through suction.

Beyond the way in which small volumes can be loaded, a way to transfer the volume displacement from the SPs is still required. A possible approach would be to fill each syringe, as well as the tubing leading up to the DTs, with the liquid of interest, although this would entail a significant dead volume. The alternative implemented here involves loading reagents with 3WVs, filling the dead volume completely with an auxiliary liquid and using it to displace the target liquid in the DTs. This liquid, referred to as the auxiliary displacement fluid here, is represented in light blue in figure 1 and has been ultrapure water in the present report. If the dilution effect at the interface with the displacement fluid is not compatible with the required lines, a significantly larger volume will be needed to occupy the entire dead volume from the SPs to the DTs (after finishing, this volume could be recovered). However, if it is handled with care and little time is allowed for diffusion, this effect will be greatly minimized.

More 3WVs can be observed immediately at the exit of the SPs, which are used to connect an auxiliary liquid reservoir (a syringe without its plunger, oriented upwards). This makes it easy to purge the system up to the DTs and eliminate potential air bubbles. Additionally, an incompressible tubing should be used between these 3WVs at the SPs' output and the 3WVs at the DTs' input, ensuring that the volume driven by the SPs matches the volume dispensed at the tips of the DTs (there should be no flexible tubing that might absorb energy). The connections can be made with various components that adhere to the Luer standard (3WVs, needles, adaptors, etc).

In the described conditions, it is possible to dispense liquid using only G-code. However, the printer is typically configured to perform **Nusteps** microsteps and move 1 mm of plastic filament (of a specific diameter) through an extruder. When dispensing liquid, it is advisable for the unit to be volumetric[12], such as µL. Thus, it is necessary to reconfigure the number of microsteps required in the stepper motor of an SP to move 1 µL. The command to be used is "M92 E*Nusteps*," where the letter E indicates that the µstep/unit change is for the Extruder (not for X, Y, or Z). To calculate this value, it is necessary to perform a gravimetric calibration of the syringe to be used in each SP. Furthermore, it is possible to determine the internal diameter of a syringe based on construction standards (BD, Becton Dickinson), which enables the calculation of the number of µsteps per unit of displaced volume according to the following equation:

$$N\mu step\ [\mu steps/\mu L] = \frac{StepsPerRev[steps/rev] \times \mu Stepping[\mu steps/step]}{\pi \times (ID[mm]/2)^2 \times lead[mm/rev]} \quad \text{Eq 1}$$

where StepsPerRev represents the number of steps per revolution in the SP stepper motor (200), µStepping is the number of microsteps per step configured in the stepper driver (16), ID denotes the inner diameter of the syringe used in the SP (BD 10 mL, 14.5 mm), and lead is the linear movement per revolution provided by the threaded rod (leadscrew) connected to the stepper (8 mm/rev). The parameters enclosed in parentheses were employed in the current research, resulting in a final value of 2.4223 µstep/µL.

As mentioned, in order to control a second SP, a new power driver and wiring are required. Furthermore, since it is common to have only one extruder, it will also be necessary to activate a second extruder in the firmware to enable multiple extrusion. In the present report, Marlin was utilized[13], and among several options, the Mixer Extruder[20] was selected. This type of extruder can melt up to 6 different plastics simultaneously and utilizes a single thermistor. This is not a trivial detail, as other configurations may require multiple thermistors, and regardless of the choice made, there must be a genuine coherence between the firmware parameters and the actual hardware connected. Since a common extruder-based printer already has a thermistor installed, there is no need for additional connections in practice. Furthermore, there are additional advantages. The mixing is activated at runtime using G-code. Otherwise, a single extruder configuration is implemented. In this way, when the original plastic extruder is reconnected, the printer resumes its original functions. Also, if instead of a mixing extruder SPs are connected, the mixing process is not carried out in practice, thus achieving the highest degree of independence for linewidths, as each SP can be utilized with a specific flow rate. Once the firmware has been updated, the command "M165 Aaa Bbb" will allow configuring a fraction "aa/(aa+bb)" of the total volume that will be dispensed by SP A, and a fraction "bb/(aa+bb)" that will be dispensed by SP B (if there were indeed a mixing procedure, those would be the fractions of the mixture). For example, if 20 µL are to be dispensed, M165 A50 B50 would yield 2 lines, each containing 10 µL, whereas M165 A75 B25 would result in 2 lines, one with A and 15 µL, and the other with B and 5 µL.

In addition to M92 and M165, there exist other significant commands. For instance, "M302 P1" enables cold extrusion, typically prohibited to prevent damage to the extruder. Since liquids are dispensed, there is no need for a minimum temperature for extruder protection (which will not even be utilized).

There are general experimental considerations that must be taken into account:
- Fixation of the NCM to the printer bed: The NCM must remain flat and securely fixed on the bed, so this glass was coated with a magnetic adhesive, and magnetic tapes were employed for the fixation of the NCM.

- Hardness of dispensing tips: the NTC is easily scratched by metal needles or rigid plastic tips, such as those used for micropipettes. In this study, the best results were achieved with catheters, which are made of inert materials and have standardized diameters. Also, in the presence of minor imperfections in the surface, catheters can flexibly adapt if necessary.
- Mounting of the DTs to the printing carriage: A coupling piece was designed, printed and attached to the effector of the 3D printer, which can be observed in figure 1 D (photograph, gray plastic) and E (diagram, in orange). As depicted, this piece holds two 3WVs (for reagent loading). Furthermore, it enables the adjustment of the angle of the DTs relative to the NCM and the line-to-line spacing.
- Vertical spacing between NCM and DTs: Different spacing tests were conducted. In some cases, there was actual spacing between NCM and DTs, while in others, DTs made contact with the NCM. The best results were achieved by slightly lowering the DTs to make contact with the NCM, and then lowering an additional 0.5 mm, causing the catheters to flex slightly.
- Bed leveling: Since the spacing and angles between DTs and NCM must remain consistent throughout the entire trajectory, bed leveling is of utmost importance. This has been documented in similar 3D printing applications, such as the direct fabrication of microstructured surfaces for planar chromatography[26].
- Coupling of SPs with DTs: As can be observed in diagram A of figure 1, SPs and DTs were interconnected utilizing 3WVs, 21G needles and Polytetrafluoroethylene (PTFE) tubing of 0.8 mm id. This type of tubing is deemed non-compressible, so it is assumed that any volume displaced from the syringe should exit the system without observable delay at the distal end.
- The utilization of dyes and BSA in preliminary test solutions: As mentioned earlier, it is advisable to carry out preliminary tests using liquids that closely resemble the final dispensing composition. Therefore, in the case of preliminary tests involving colored solutions, a final concentration of 1% BSA was also employed.
- Additional Features of the proposed setup: There are features that were not utilized in this study. For example, the 3D printer bed also allows for temperature control. Since this study was designed for the dispensing of biological fluids that are typically sensitive even to low temperatures (below 40 °C), this feature was not utilized here. However, for other types of dispensing, such as Enzyme-Linked Immuno Sorbent Assay (ELISA) plate reactions, this or other functions could prove to be useful. Furthermore, it's worth highlighting that in the suggested configuration, the 3D printer maintains its ability to execute 3D movements, which means its application extends beyond LFIA lines and can be adapted for other microfluidic paper-based devices, whether featuring 2D or 3D geometries[27]. And, in general terms of dispensing (not only on paper or NCM), there is regulable volumetric control.
- Regarding several of the required components, the proposed method utilizes standardized and readily available nursing/medical supplies, such as 3WVs, catheters, syringes, needles. In turn, these components have Luer inputs/outputs, which also standardizes the connection between all of them. Additionally, in classic secondary medication sets, there is usually an extension of the tubing that ends in a male Luer connector. Two of these, coupled with the same tubing, make up a male-male Luer adapter. With these adapters, and the 3WVs, it is possible to make any combination of fluid connections. The use of these components enhances the potential for replicating the proposed setup in various laboratories. This also implied the printing of open-source syringe pumps based on NEMA motors, classic 3D printing rods, among others, as well as the adaptation of a commercial 3D printer with open-source firmware, which is currently present in the infrastructure of many laboratories.

### 3.2. Single dispensing: Dispensing Speed and Dispensing Rate

In the experiments described below, sequential dispensing was carried out with a single SP, thus they can be performed without firmware modifications. If the power supplied by the extruder driver is sufficient for the SP, and if the extruder wiring can handle this energy, it is only necessary to connect the SP and use runtime code. A simple "M92 Exx" command will suffice to reconfigure the number of µSteps/µL, a parameter that, in these experiments, was set to 2.4223. The experiments were conducted in the same NCM, but in one case with different DS (Dispensing Speed, linear displacement speed of the NCM, mm/min) and the same dispensed volume, and in the other case, they were performed with the same DS and variable dispensed volumes. This resulted in different DR (Dispensing Rate, µL/mm). A consistent distance of 200 mm was traversed, encompassing a portion over the NCM. As in previous reports[16], before image analysis, the initial part of every dispensing (40 mm) was excluded due to initial excesses (stains from the first contact of the NCM with the DT) and subsequent narrowing before reaching a Steady State Zone. Subsequently, 70 mm of NCM were analyzed, obtaining the average width of each line every 2.5 mm (28 bins). In figure 2, results obtained from the analysis of these NCM regions can be observed.

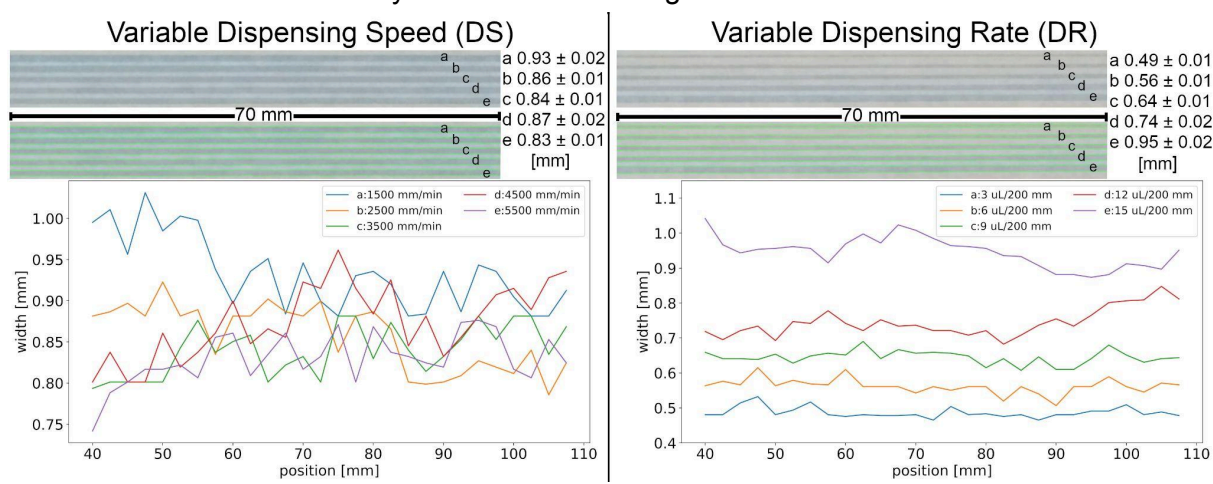

Figure 2: Line widths for single dispensing with variable DS or DR. Left: variable DS (always with 15 µL). Right: variable DR (always at 1500 mm/min). Up: Photographs, detected edges (green) and CI for line widths in mm. Down: Width values from detected edges every 2.5 mm. All solutions had 1% BSA and blue dye.

On the left side of figure 2, it can be observed that, despite some differences, in the steady state, especially from approximately 60 mm, the lines closely resemble each other, and the widths obtained were in the range of classic linewidths for immunoassays[28]. Before 60 mm, the width of line 1 (1500 mm/min) is noticeably greater. To ensure that the DT is filled before starting to dispense on an NCM, all lines have an initial priming outside the membrane. At the beginning of the dispense of line 1, there was a higher accumulation of liquid (additional manual priming). This excess justifies the greater width, especially through the effect referred to here as the "brush" or "drag" effect. In other words, beyond the internal content that is flowing and will be dispensed by the DT, if there is external content, it will be dragged until it is eventually depleted, ultimately reaching a steady state (only internal flow). Excluding line 1 for the reasons mentioned and with the exception of line 4 (4500 mm/min), in which widening in the middle section (approximately 70-80 mm) was detected, the average width of the remaining lines is not different according to the ANOVA test.

The DT used (22G catheter) has an outer diameter of 0.9 mm. It should be noted that this value was only exceeded on a few occasions, including when the "brush" effect is present, and it is very close to the average width of the lines. This implies that if there are no flow excesses or defects, in addition to the internal width of a DT, its external width will be one of the determining factors in the obtained line width. It is possible to obtain wider lines with higher flows and the "brush" effect, but it is reasonable to assume that beyond a certain flow rate, the "brush" effect may not be sufficient to achieve a uniform line width, leading to defects such as significant accumulations on the lines.

The similarity in the dispensed line widths at different DS (Dispensing Speed, linear displacement speed of the NCM, mm/min) arises from a particular aspect of the proposed setup. In this work, increasing DS does not necessarily lead to a decrease in the dispensed line width, as has been reported in similar studies[16] and as one might intuitively expect. Since the 3D printer has control over both the bed (Y-axis, dispensing direction) and the SP, it is possible to use G-code for simultaneous extrusion and movement (G1, as G0 is reserved for non-extrusion movements). For example, the G-code "G1 F3000 Y200 E20" will execute a 200 mm movement at a bed speed (DS) of 3000 mm/min while proportionally dispensing 20 µL. This implies, on one hand, that the parameter DR (Dispensing Rate, µL/mm) will take the value of 0.1 µL/mm (20 µL/200 mm), and on the other, it's crucial to note that the stepper speed of the SP is not set by the user but is instead recalculated by the firmware of the printer to traverse 200 mm and dispense proportionally 20 µL. In other words, if speed limits imposed by the firmware are not exceeded, the flowrate of a SP will depend on the total dispensed volume, the distance to travel, and the DS, as expressed by the following equation:

$$SP_{flowRate} = \frac{totalVolumeDispensed[\mu L] \times DS[mm \cdot min^{-1}]}{totalDistance[mm]} = DR \times DS \quad \textbf{Eq. 2}$$

Within certain limits of DS (in this case, between 1500 and 5500 mm/min), the value used may be more empirical, for instance, to achieve good dispensing in a short time, but it may not significantly affect the linewidths. Also, the value of DR will provide direct information regarding the volume of liquid required to achieve a certain total line length. Furthermore, because of its relation with concentration and mass, the parameter of volume per unit length appears to be a better descriptor in relation to the obtained width (and, consequently, the intensity in the signal of the test) than the DS and the flowrate at which it was obtained.

The analysis of the curves on the right side of figure 2 indicates that the curves with total volumes of 3 to 9 µL have equivalent variances (Bartlett test), but their widths are not statistically similar (ANOVA test). On the other hand, the curves for 12 and 15 µL also have equivalent variances, and their widths are not statistically similar (t-test). All of these differences are visually apparent and were expected. It should be noted that line 1 (DS = 1500 mm/min, 15 µL total) in the left-hand curves has the same parameters as line 5 in the right-hand curves (15 µL total, DS = 1500 mm/min). If a statistical analysis is performed, both curves result homoscedastic and show no significant differences in linewidth according to the t-test.

3.3. Simultaneous dispensing

As mentioned, the dual and simultaneous dispensing experiments were conducted by independently managing two SPs, for which it was necessary to modify the firmware once to enable the Mixing Extruder mode. Two experiments were conducted and two videos were recorded. An image analysis was carried out on a snapshot from the second video, following a similar procedure as the one previously done. The same DTs as in previous experiments were used, with a DS of 3000 mm/min and a 7 mm separation between DTs. Solutions of 1% BSA with blue and orange dyes were dispensed. The NCM used was small, as one extra goal was to observe the behavior of the SPs outside of the NCM. This information is valuable for assessing whether, under the configured conditions, there is an excess or lack of liquid, or if there are issues related to the SPs themselves rather than the NCMs. The total travel distance was 180 mm. As before, the beginning of the NCM was not considered for line detection and subsequent statistical analysis (only in the steady-state zone). In both cases, 24 µL of liquids were dispensed. Results can be seen in figure 3.

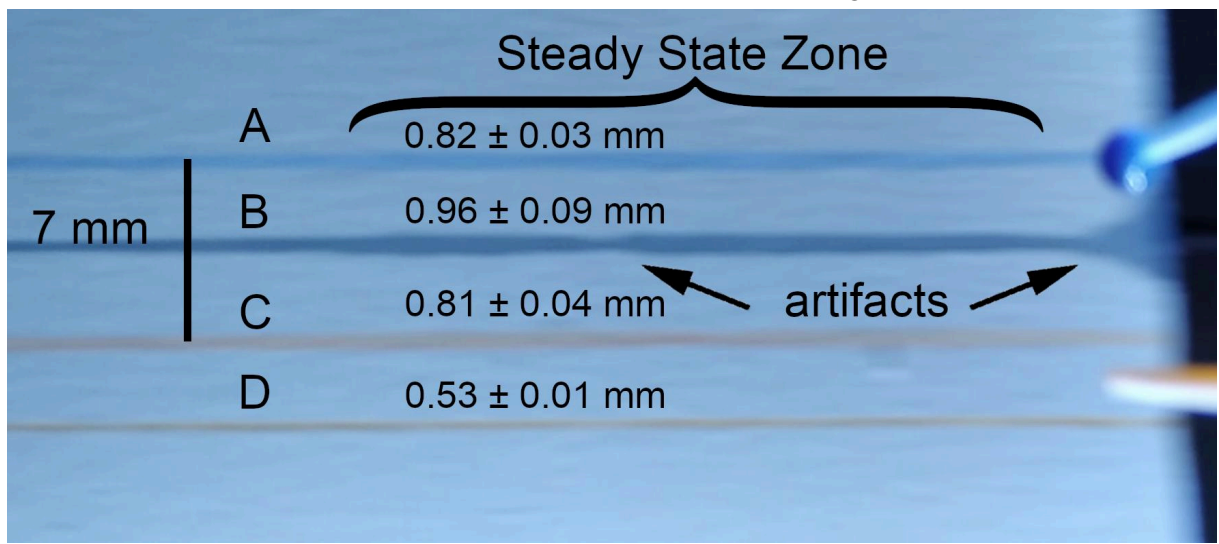

Figure 3: Simultaneous dispensing: 24 µL of colored liquids were dispensed simultaneously by 2 SPs over a 180 mm travel distance. First dispensing, lines A and C: 12 µL each. Second dispensing, lines B and D: 19.2 µL and 4.8 µL, respectively. Confidence intervals were obtained by statistically analyzing only the steady-state zone.

In the first simultaneous dispensing (lines A and C), each SP dispensed 50% of the volume (i.e. "M165 A50 B50" before dispensing). Figure 3 shows that both linewidths were similar, which is consistent with the equality of volumes and DTs. Each line was dispensed with an approximate DR of 66.7 pL/mm (12 µL/180 mm). If results of experiments with single dispensing and variable DR (figure 2, right side) are evaluated, it can be observed that the lines with widths of 0.74 mm and 0.95 mm had DRs of 60 and 75 pL/mm, respectively. Therefore, the obtained widths (0.82 and 0.81 mm) are consistent with a DR of 66.7 pL/mm. Additionally, the experiments were conducted at different DS (1500 and 3000 mm/min), which also had no significant effects.

Subsequently, the DTs were repositioned, and lines B and D were obtained, representing 80% and 20% of the total 24 µL (i.e. "M165 A80 B20" before dispensing). The dispensed volumes were 19.2 µL and 4.8 µL, respectively, over 180 mm, resulting in DRs of 26.67 pL/mm and 106.67 pL/mm, respectively. As expected for line D, the width obtained for 15% of the volume was smaller. Its DR of 26.67 pL/mm and its width of 0.53 mm were consistent with the values obtained in the figure 2 (right side), for the dispensed volumes of 3 and 6 µL over 200 mm (0.487 mm with 15 pL/mm and 0.562 mm with 30 pL/mm, respectively). Regarding line B, the resulting DR of 106.67 pL/mm is higher than all previously used values. While it also yielded the widest line (0.96 mm), one might have expected a wider width for this DR. However, based on the previous experiments, it is known that with this DT, the maximum achievable widths were close to 0.96 mm. This constraint implies an upper limit on the width, and therefore, seeding with a DR of 106.67 pL/mm is excessive under these conditions. This implies that the liquid does not get absorbed uniformly into the NCM, as evidenced by the observed artifacts and confirmed by a wider confidence interval around the mean width of 0.96 mm.

Regarding the dispensed lines outside of the NCM, in the first case (50% and 50%), continuous lines with a constant width were observed, without excesses or interruptions. On the other hand, in the second case (80% and 20%), for the thin line, there were discontinuities not caused by SP pulses but rather due to a very low DR. For the thick line, a messier exit from the NCM was apparent compared to the rest, with a continuous line without pulses, appearing not so much wider but rather taller/thicker, in line with the excess in DR.

3.4. Study case

As an illustrative application, the dispenser was employed to deposit the reagents for developing a LFIA capable of detecting specific IgG antibodies for human leptospirosis. Similar conditions to those used in section 3.3 were employed. For a 150 mm section of NCM, 10 µL of both reagents (TL and CL) were dispensed while keeping the same DT, DS, and DR. Figure 4 presents a simplified schematic of the membrane assembly used in a lateral flow test for human leptospirosis detection (A). The schematic includes the following key components:

- Nitrocellulose Membrane: A thin strip of nitrocellulose membrane where the Test and Control lines were deposited.
- Test Line: The region on the membrane where the antigen was immobilized, corresponding to *Leptospira interrogans* homogenate at a concentration of 0.5 mg/mL.
- Control Line: An area adjacent to the test line that contains anti-IgG goat antibody at a concentration of 1 mg/mL. This zone captures the labeled conjugate and serves as a validation of the test's integrity.
- Conjugate Pad: A reddish-colored area where gold particles conjugated with anti-human IgG goat antibodies are deposited.
- Sample Pad: The area where the sample is applied, consisting of 1/10 diluted positive and negative control serum in the sample buffer.
- Absorbent Pad: The area where assay fluids are retained.

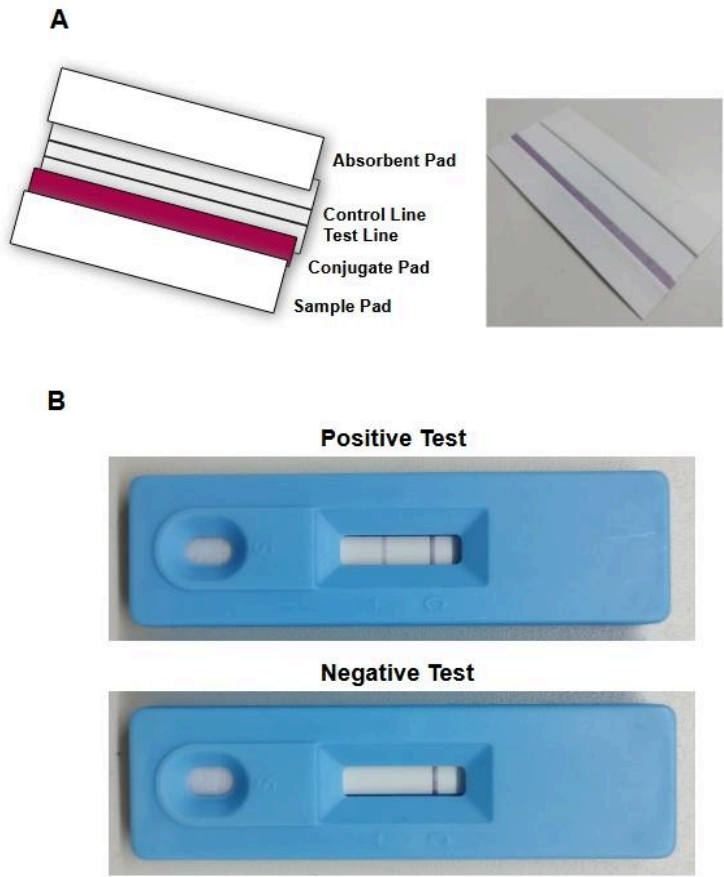

Figure 4: Development of an LFIA for the detection of human leptospirosis. A: Schematic representation and a photograph of the LFIA structure. B: Results obtained from assessing the tests using positive and negative control samples.

On the right, a photograph of the actual assembly is shown before being cross-cut with a guillotine to obtain 4mm-wide reactive strips. It should be noted that once the biomolecule-seeded lines have dried, they are no longer visible. These strips are subsequently placed in the housing (B). Following the application of 100 uL of positive and negative samples in the designated area, it starts migrating across the membranes, carrying the conjugate and reacting with the specific antibodies present in the positive sample. In this case, both lines are visible. Conversely, in the absence of specific antibodies in the sample, only the control line is observed.

The proposed configuration allowed the development of an LFIA for Leptospirosis detection and it also offers additional advantages. Firstly, in the infrastructure of a contemporary laboratory, it is customary (or advisable) to have a 3D printer. The cost of this equipment is low, much lower than several other laboratory instruments, including commercial SPs. Furthermore, the components used in this work are also inexpensive. Whether they are 3D printing supplies for producing SPs (plastic filaments, power drivers, NEMA steppers, linear bearings) or nursing/medical standards, they are readily available on a global scale. Secondly, everything is controlled by the 3D printer, eliminating both the need to render it unusable for plastic printing and the requirement for manual actuator coordination, even for simultaneous dispensing with different linewidths. Furthermore, while it was demonstrated as an example for LFIA, the proposed setup is not limited to that, as it possesses additional capabilities (temperature control, 3D positioning, general adjustable volumetric control, among others). Thirdly, whereas other setups necessitate filling the entire dead volume with reagents or, conversely, are constrained to small loads (only within the DT), the one proposed here can be adapted to any situation.

# 4. Conclusions

The successful dispensing of reagents for immunochromatographic test strips was achieved through the integration of 3D printers, syringe pumps and nursing/medical basic supplies. With the setup reported, it was possible to actively dispense both individual and simultaneous lines. The unified control of the 3D printer enables dual, active, regulable and simultaneous dispensing, four features that are typically found only in certain high-cost commercial equipment. In all cases the width of these lines could be configured within certain empirical limits. As a practical example, a genuine LFIA for leptospirosis was also obtained. Also, the setup can be adapted to the volume actually available for dispensing.
All the resources used are low cost and globally accessible, which facilitates the reproducibility of the setup. This includes the nursing supplies, the SPs built, the 3D printer itself and also the firmware used (open source and freely available). The standard function of the printer can be restored after reconnecting the extruder, and in addition, it can be used to perform other tasks not exemplified here (other types of dispensing, such as for non-linear geometries paper-based devices or in ELISA plates, temperature control, among others).

# Acknowledgments

We are grateful to Agencia Nacional de Promoción Científica y Tecnológica (ANPCyT): PICT 2017-2040, Consejo Nacional de Investigaciones Científicas y Técnicas (CONICET): PIBAA 28720210100763CO and Universidad Nacional del Litoral (UNL): CATT Nº: 06-01-2021 for their financial support. And to Dra. Bibiana Vanasco (Instituto Nacional de Enfermedades Respiratorias "Dr. Emilio Coni", Santa Fe, Argentina) for the provision of the sera samples.